\documentclass[conference]{IEEEtran}
\usepackage{cite}
\usepackage{amsmath,amssymb,amsfonts}
\usepackage{algorithmic}
\usepackage{graphicx}
\usepackage{textcomp}
\usepackage{xcolor}
\usepackage{multirow}
\usepackage{array}
\def\BibTeX{{\rm B\kern-.05em{\sc i\kern-.025em b}\kern-.08em
    T\kern-.1667em\lower.7ex\hbox{E}\kern-.125emX}}
\begin{document}

\title{Developing Imperceptible Adversarial Patches to Camouflage Military Assets From Computer Vision Enabled Technologies}

\author{\IEEEauthorblockN{Chris Wise, Jo Plested}
\IEEEauthorblockA{\textit{School of Engineering and Information Technology} \\
\textit{University of New South Wales}\\
Canberra, Australia \\
c.wise@student.unsw.edu.au, j.plested@unsw.edu.au}}

\maketitle

\begin{abstract}
Convolutional neural networks (CNNs) have demonstrated rapid progress and a high level of success in object detection. However, recent evidence has highlighted their vulnerability to adversarial attacks. These attacks are calculated image perturbations or adversarial patches that result in object misclassification or detection suppression. Traditional camouflage methods are impractical when applied to disguise aircraft and other large mobile assets from autonomous detection in intelligence, surveillance and reconnaissance technologies and fifth generation missiles. In this paper we present a unique method that produces imperceptible patches capable of camouflaging large military assets from computer vision-enabled technologies. We developed these patches by maximising object detection loss whilst limiting the patch's colour perceptibility. This work also aims to further the understanding of adversarial examples and their effects on object detection algorithms.
\end{abstract}


\section{Introduction \label{sec:Introduction}}
Convolutional neural networks (CNNs) have been shown to be effective at object detection \cite{krizhevsky2012imagenet, girshick2014rich, li2020deep, masi2018deep, mazurowski2019deep}. However, it has been demonstrated that CNNs can be successfully targeted and disrupted through designed attacks or adversarial patches \cite{ goodfellow2014explaining, szegedy2013intriguing, su2019one, brown2017adversarial, moosavi2017universal}. As modern CNNs are increasingly applied across various domains \cite{rawat2017deep, cheung2020graph, gu2018recent, pinheiro2014recurrent, mazurowski2019deep}, it is important to understand these CNNs and their vulnerabilities. Specific to a military context, these designed attacks or adversarial patches, when printed, could be used to attack an adversaries' CNN detectors and could act as passive defence against object detectors. This need to camouflage from object detectors is vital as intelligence, surveillance and reconnaissance (ISR) technologies, and fifth generation missiles integrate vision technologies \cite{wang2021object, kechagias20163d, zhu2018visdrone, berker2017feature, belmonte2019computer}. Traditional camouflage methods are impractical for larger mobile assets, such as aircraft, so adversarial patches should be further developed as an alternative camouflage technique. Our proposed method for the development of adversarial patches is unique and designed to disguise their presence to humans whilst still being capable of acting as an alternative camouflage against detection algorithms. Our method was developed by combining adversarial techniques that focus on patch deception strength and imperceptibility. Specifically, we make the following contributions:
\begin{enumerate}
\item a discussion of the concerns and constraints of a military using adversarial patches to camouflage large mobile objects 
\item a unique algorithm that develops imperceptible, effective and robust adversarial patches 
\item recommendations for future research into adversarial patch development.
\end{enumerate}
In Section \ref{sec:Related_Work}, we introduce adversarial patches by reviewing relevant current and foundational literature. Algorithm and experiment methodologies are described in Section \ref{Methodology}, with Section \ref{sec:Results} detailing experiment results. Section \ref{Discussion} considers research results and experiment validity, with the broad positive and negative impacts of the research being considered in Section \ref{Broader_Impact}. The paper concludes with Section \ref{Conclusions} and finishes with Section \ref{Future_Work}, which provides suggestions for future adversarial patch development.
\section{Related Work \label{sec:Related_Work}}
Researchers have produced imperceptible adversarial methods that disrupt computer vision algorithms, such as CNNs \cite{goodfellow2014explaining, szegedy2013intriguing}, with even one-pixel attacks shown to be effective \cite{su2019one}. Early attacks were often specific to single images and lacked portability \cite{moosavi2017universal}. However, later research identified an algorithm that generates more universal patches \cite{brown2017adversarial}.  Specifically, a patch application operator \(A(p,x,l,t)\) was used to create universal and robust patches. Here, \(p\) is a patch transformed (rotated and scaled) by \(t\) and then applied to the example image \(x\) at location \(l\). The patches generated from this operator were suitable for real-world attacks where patches appear in various backgrounds, locations, and lighting conditions. A patch \(\hat{p}\) is then generated by optimising the objective function \cite{brown2017adversarial}: 
\[\hat{p} = arg \mathop{max}_{\textbf{p}} \mathbb{E}_{x{\sim}X,t{\sim}T,l{\sim}L} [log  {Pr(\hat{y}|A(p,x,l,t))}]\]
Where \(X\) is the set of training images, \(T\) is a distribution of patch transformations, and \(L\) is a distribution of patch locations. This patch application operator can be configured to camouflage patches by adding \(|| p - p_{orig}||_{\infty} < \epsilon\) where \(p_{orig}\) is some starting desired disguise. When using this constraint, the trained patch is successful but not as effective as its undisguised counterpart \cite{brown2017adversarial}.

\begin{figure}[htbp]
\centerline{\includegraphics[width=0.4\textwidth]{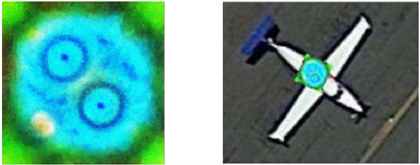}}
\caption{Example Patch and Placement on an Aircraft \cite{den2020adversarial}}
\label{salient_patch_and_aircraft}
\end{figure}

Researchers have examined using adversarial patches to disguise assets from automatic object detection \cite{aurdal2019adversarial, wu2020making, sharif2016accessorize}. One specific application used adversarial patches to disguise military aircraft from object detectors trained on drone surveillance footage \cite{den2020adversarial}. This work showed that adversarial patches could be used as an alternative camouflage against automatic object detectors and uniquely applied and studied the efficacy of different adversarial patch hyperparameters. Specifically, explored were different patch locations and sizes relative to the military assets and the saliency and number of patches. However, the patches studied were notably conspicuous and  humanly perceptible—an example of a salient patch is seen in Fig.~\ref{salient_patch_and_aircraft}, and thus unlikely to be used in a military context. Further, as Fig.~\ref{salient_patch_and_aircraft} highlights, the studied patches were unrealistic in size and covered a significant portion of the aircraft. Consequently, these larger patches sat outside object boundaries and could not be printed and placed on an aircraft.

Other research has continued to improve disguising adversarial patches whilst maintaining their robustness \cite{carlini2017towards, papernot2016limitations, rony2019decoupling}. Recent work \cite{zhao2020towards} has challenged the assumption that imperceptibility is achieved using the constraints proposed in earlier literature \cite{brown2017adversarial, goodfellow2014explaining}. This research created imperceptible patches by minimising the patch’s perceptual colour (PerC) distance \cite{luo2001development} whilst training its adversarial strength using the approach proposed by Carlini \& Wagner (C\&W) \cite{carlini2017towards}. The researchers found that combining the PerC distance loss with the C\&W approach (PerC-C\&W) can produce robust but imperceptible adversarial patches. The approach was further improved by alternating between PerC distance and detection loss updates (PerC-AL). Though PerC-C\&W and PerC-AL can produce robust imperceptible image perturbations, the authors prefer PerC-AL as the algorithm reduces hyperparameter optimisation time. However, this study explored only image perturbations with the same dimension as the target image. Thus, the existing PerC-AL and PerC-C\&W algorithms cannot be used for example to create a  patch only located within the bounds of a target object in the image. Printing and applying PerC-AL and PerC-C\&W patches to real-world objects remains unexplored.
\newline
\begin{figure}[htbp]
\centerline{\includegraphics[width=0.48\textwidth]{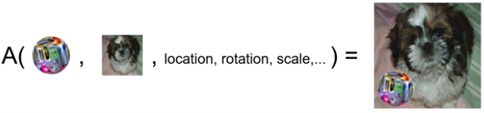}}
\caption{An illustration of the patch application operator \cite{brown2017adversarial}}
\label{google_adversarial_patch_operator}
\end{figure}
\newpage

Further work has also developed frameworks for creating more robust physical adversarial patches \cite{eykholt2018robust, sharif2019general, thys2019fooling, chen2018shapeshifter, bose2018adversarial, liu2018dpatch}. Recent research has improved upon previous attacks by creating perturbations that do not need to occlude targeted objects, and yet such targeted objects are still misclassified or not detected \cite{lee2019physical}. Specifically, researchers developed a patch capable of suppressing detections even for objects at a distance from the patch. Consequently, a user does not need to manipulate the target object actively. These patches were developed by maximising the loss of an object detector using the fast sign gradient method (FGSM) \cite{goodfellow2014explaining}. The patch was clipped into RGB colour space whilst training and therefore was printable once developed. This specific patch development implementation has become known as Robust-DPatch \cite{artTrustedAI}. However, the authors did not explore if Robust-DPatch maintains its robustness when made less perceptible.
\section{Methodology\label{Methodology}}
The Adversarial Robustness Toolbox (ART) V1.8 library’s \cite{nicolae2018adversarial, artTrustedAI} Robust-DPatch \cite{lee2019physical}, and Faster R-CNN wrapper \cite{ren2015faster} implementations were adapted for these experiments. Most significantly, perceptibility updates based on \cite{zhao2020towards} were integrated. The implemented code is available here: \textcolor{blue}{https://github.com/ChrisWise07/imperceptible-patch-generator}

\begin{figure}[htbp]
\centerline{\includegraphics[width=0.28\textwidth]{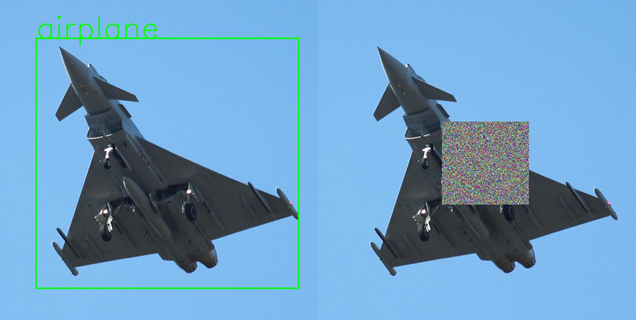}}
\caption{Bounding Box Setting Patch Location and Size}
\label{example_patch_with_bound_box}
\end{figure}

\begin{figure}[htbp]
\centerline{\includegraphics[width=0.11\textwidth]{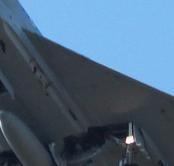}}
\caption{Cropped Segment of Image}
\label{example_image_segment}
\end{figure}

Five specific images were chosen from the COCO-2017 dataset \cite{lin2014microsoft}. The chosen images represent some typical situations in which a plane could be found, grounded, flying (with \& without cloud coverage), taking off, and at a distance. As Fig.~\ref{example_patch_with_bound_box} shows, initial Faster R-CNN prediction bounding boxes were then used to set the patch location to the box’s centre and scale the patch shape to a ratio of this box. The bounding box was used to set patch size and location to help the patch sit within object bounds. Then as Fig.~\ref{example_image_segment} demonstrates, the original image's corresponding patch shape and location were cropped out and used for PerC distance comparison calculations. Using only the image segment for PerC distance calculations was preferred as early experiments indicated this method was faster and did not hinder perceptibility performance compared to perceptibility calculations that used the whole image or the image segment contained within the bounding box for PerC. 

A patch specific to each image was developed, following \cite{lee2019physical}, by updating the patch to maximise prediction loss whilst clipping it to within RGB space to maintain its printability. During training, before the object detector performed a prediction, an image was first transformed by rotating the image by [0, 90, 270]°. These rotations were chosen as they transform an aircraft into realistic situations, that is, the aircraft being in steep ascent or descent or flying level. Borrowing from \cite{lee2019physical}, the image brightness was also adjusted brightness by factor from [0.4,  1.6] and cropped at the edges so that the patch occupied 20\%-30\% of the image. Then techniques from \cite{zhao2020towards} were implemented which allowed the algorithm to uniquely update the patch to minimise its perceptibility by minimising the PerC distance between it and the corresponding image segment. Finally, a mechanism was added to expand upon and generalise \cite{zhao2020towards}, which allowed only alternating deception and perceptibility updates. This was achieved by implementing a further hyperparameter that controls the ratio of perceptibility to deception updates by performing a deception update when:
\[i \pmod n = 0\]
Where $i$ is the current iteration number, and $n$ is frequency hyperparameter.

Initial hyperparameters chosen matched those found in \cite{lee2019physical, zhao2020towards}, with these hyperparameters made explicit in Table~\ref{experiment_hyperparameters}. However, hyperparameters were optimised in ablation studies of 50 steps, where a step is defined as 1000 iterations \cite{lee2019physical}. When optimising the perceptibility to deception ratio, the number of iterations per step was changed so that 1000 deception updates were still performed at each step. In every nth step, the deception learning rate (DLR) was decayed by 0.95, whereas the perceptibility learning rate (PLR) was decayed from a maximum to 0.1 of this maximum using cosine annealing. Subsequent experiments assessed decaying the maximum PLR at the same rate as the DLR.   

The experiments ran on compute nodes containing an Nvidia V100 GPU and 12 cores from one processor. Running time varied but approximately required 11 hours per ablation study. The final experiment was conducted on the same computational resources for 100 steps and executed for approximately 20 hours.
\section{Results \label{sec:Results}}
The final optimised \textbf{imperceptible patch} achieved \textbf{6.71 mAP-50} (mAP at 0.5 IOU) with a \textbf{2854.93 PerC distance}, making imperceptible patch more imperceptible than, whilst being as robust as, \textbf{Robust-DPatch}, which achieved \textbf{7.27 mAP-50} with a \textbf{4346.61 PerC distance}. The visual differences between imperceptible patch and Robust-DPatch are seen in Table \ref{final_dpatch_vs_imperceptible_patch}. Note that the the reported mAP concerns only aircraft detections and is averaged across different object-confidence thresholds, with the thresholds of [0.5, 0.1, 0.001] being used. 

The conservative 0.5 IOU and object-confidence thresholds were chosen to meet military concerns. That is, it is reasonable to foresee a military using conservative benchmarks for true positive detection as a partially correct detection still poses a risk to asset or human life. Thus, militaries will desire a patch capable of suppressing conservative detections.

The following observations and conclusions were made from the ablation study results displayed in Table~\ref{ablation_study_results}:
\begin{itemize}
    \item Results support a relationship between patch perceptibility and robustness. This relationship is demonstrated in Studies 5 \& 8, where decreased mAP (increased robustness) is correlated with increased perceptibility.
    \item Based on the evidence from Experiment 3, an optimal DLR balances extreme updates. That is, the learning rate is large enough to be effective but small enough to avoid extreme and subsequently perceptible updates.
    \item Studies 9 \& 10 evidence that patch perceptibility is a more complex space that is harder to optimise and was being dominated in earlier ablation studies by deception updates. Explicitly, Study 9 running for 50 steps, found the optimal configuration decays the DLR every 5$^{th}$ step. However, the longer study of 100 steps found that the optimal configuration should decay the maximum PLR as well, and without maximum PLR decay, the patch becomes weakly adversarial. This demonstrates that perceptibility updates could only dominant once the DLR was small enough.
    \item Study 4 supports \cite{lee2019physical}'s use of momentum for deception updates and Study 6 shows that momentum for perceptibility updates also leads to greater patch performance.
    \item The results from Study 2 confer with \cite{lee2019physical, brown2017adversarial, den2020adversarial, liu2018dpatch} and show that transformations result in more robust patches.
\end{itemize}

Points of note for correct interpretation of results:
\begin{itemize}
    \item The score displayed is the addition of the rank of an experiment's mAP and average patch PerC distance within a particular study, with the objective being to achieve the lowest mAP, average patch PerC distance and subsequently overall score.
    \item The lowest possible score is 2 for each experiment.
    \item Within the tables, the best mAPs, PerC distances, and overall scores within the tables are bolded.
    \item The optimal hyperparameter for an ablation study is also bolden. 
    \item If a hyperparameter was found to be optimal, all following ablation studies updated the corresponding default hyperparameter to that optimal configuration. All resulting optimal configurations were used to develop the imperceptible patch utilised in the final comparison study. Note that these optimal hyperparameters are made explicit in Table~\ref{experiment_hyperparameters}. 
    \item If two or more hyperparameters shared the same score, the study that executed the quickest was chosen. If execution time was equal, than the hyperparameter that best balanced robustness and imperceptibility was chosen.
    \item The hybrid initial patch configuration used a cropped image segment with RGB colour noise added to it.
\end{itemize}
\begin{table*}[t]
\centering
\caption{Ablation Study Results}
\begin{tabular}{|c|c|c|c|c|c|}
\hline
\textbf{Study Number} & \textbf{Experiment Focus} & \textbf{Hyperparameter} & \textbf{mAP (\%)} & \textbf{Mean PerC Distance} & \textbf{Combined Score}\\
\hline
\multirow{4}{*}{1} & \multirow{4}{*}{Initial Patch configuration} & Black & 20.93 & 3683.05 & 7 \\
&& \textbf{Hybrid} & \textbf{12.8} & 2672.39 & \textbf{3} \\
&& Image Segment & 63.18 & \textbf{2633.06} & 5 \\
&& Random & 14.06 & 3150.76 & 5 \\
\hline
\multirow{3}{*}{2} &
\multirow{3}{*}{Transformations} & No transformations & 43.54 & 3726.7 & 6 \\
&& \textbf{Deception prediction only} & 12.8 & \textbf{2672.39} & \textbf{3} \\
&& Deception \& perceptibility predictions & \textbf{9.73} & 3417.96 & 3 \\
\hline
\multirow{3}{*}{3} &
\multirow{3}{*}{Deception Learning Rate} & 0.05 & 28.39 & \textbf{2352.15} & 4 \\
&& \textbf{0.1} & \textbf{12.8} & 2672.39 & \textbf{3} \\
&& 0.2 & 14.71 & 3248.64 & 5 \\
\hline
\multirow{3}{*}{4} &
\multirow{3}{*}{Deception Momentum Amount} & 0.8 & 18 & 2675.71 & 5 \\
&& \textbf{0.9} & \textbf{12.8} & 2672.39 & \textbf{3} \\
&& 0.95 & 31.95 & \textbf{2663.88} & 4 \\
\hline
\multirow{3}{*}{5} &
\multirow{3}{*}{Max Perceptibility Learning Rate} & 0.25 & \textbf{9.01} & 3202.3 & 4 \\
&& \textbf{0.5} & 12.8 & 2672.39 & \textbf{4} \\
&& 1.0 & 33.99 & \textbf{1827.12} & 4 \\
\hline
\multirow{3}{*}{6} &
\multirow{3}{*}{Perceptibility Momentum Amount} & 0.8 & 16.22 & 2662.65 & 6 \\
&& \textbf{0.9} & \textbf{10.88} & \textbf{2635.27} & \textbf{2} \\
&& 0.95 & 12.47 & 2636.45 & 4 \\
\hline
\multirow{4}{*}{7} &
\multirow{4}{*}{Momentum Use} & No momentum & 28.34 & 2645.26 & 6 \\
&& Deception only momentum & 12.8 & 2672.39 & 7 \\
&& Perceptibility only momentum & 12.18 & 2667.64 & 5 \\
&& \textbf{Deception \& Perceptibility momentum} & \textbf{10.88} & \textbf{2635.27} & \textbf{2} \\
\hline
\multirow{3}{*}{8} &
\multirow{3}{*}{Deception Frequency Controller ($n$)} & \textbf{1} & \textbf{10.88} & 2635.27 & \textbf{4} \\
&& 2 & 36.39 & 1809.23 & 4 \\
&& 4 & 67.56 & \textbf{854.44} & 4 \\
\hline
\multirow{4}{*}{9} &
\multirow{4}{*}{Decay Rates} & \textbf{5} & \textbf{10.88} & \textbf{2635.27} & \textbf{2} \\
&& 5$^{\mathrm{a}}$ & 13.62 & 2952.68 & 6 \\
&& 10 & 15.05 & 2794.34 & 5 \\
& &10$^{\mathrm{a}}$ & 15.1 & 2909.16 & 7 \\
\hline
\multirow{4}{*}{10} &
\multirow{4}{*}{Decay Rates$^{\mathrm{b}}$} & 5 & 39.52 & \textbf{2118.66} & 5 \\
&& \textbf{5$^{\mathrm{a}}$} & \textbf{6.71} & 2854.93 & \textbf{4} \\
&& 10 & 30.61 & 2483.14 & 5 \\
&& 10$^{\mathrm{a}}$ & 18.16 & 2859.36 & 6 \\
\hline
\multicolumn{5}{l}{$^{\mathrm{a}}$with max perceptbility learning rate decay, $^{\mathrm{b}}$executed for 100 steps}
\end{tabular}
\label{ablation_study_results}
\end{table*}
\begin{table*}[t]
\caption{Comparison Study Results}
\begin{center}
\begin{tabular}{|c|c|c|c|}
\hline
\textbf{Patch Type}&\textbf{mAP (\%)}&\textbf{Mean PerC Distance}&\textbf{Combined Score}\\
\hline
No Patch & 99.99 & 0 & 6 \\
\hline
White Patch & 68.64 & 5312.76 & 7 \\
\hline
Black Patch & 69.54 & 5872.89 & 9 \\ 
\hline
Robust DPatch & 7.27 & 4346.61 & 5 \\
\hline
\textbf{Imperceptible Patch} & \textbf{6.71} & \textbf{2854.93} & \textbf{3} \\
\hline
\end{tabular}
\label{final_comparison_study}
\end{center}
\end{table*}
\begin{table*}[t]
\caption{Visual Comparison Between Robust-DPatch and Imperceptible Patch}
\begin{center}
\begin{tabular}{|m{2cm}<{\centering}|m{6cm}<{\centering}|m{6cm}<{\centering}|}
\hline
&\textbf{Robust-Dpatch$^{\mathrm{a}}$}&\textbf{Imperceptible Patch (ours)}\\
\hline
Generated Patch & \includegraphics[width=0.3\textwidth]{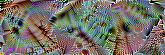} & \includegraphics[width=0.3\textwidth]{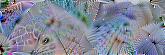} \\
\hline
Image augmented with Patch & \includegraphics[width=0.3\textwidth]{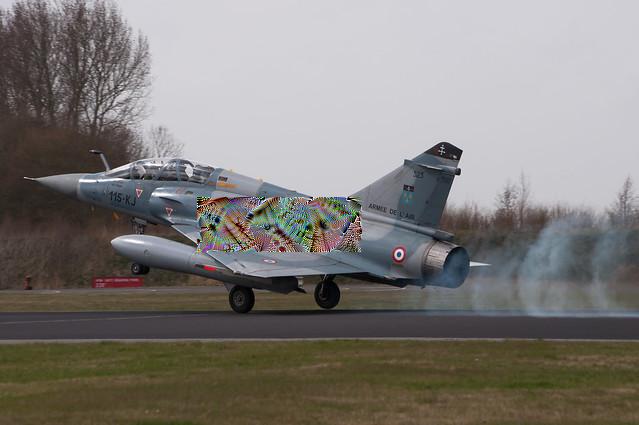} & \includegraphics[width=0.3\textwidth]{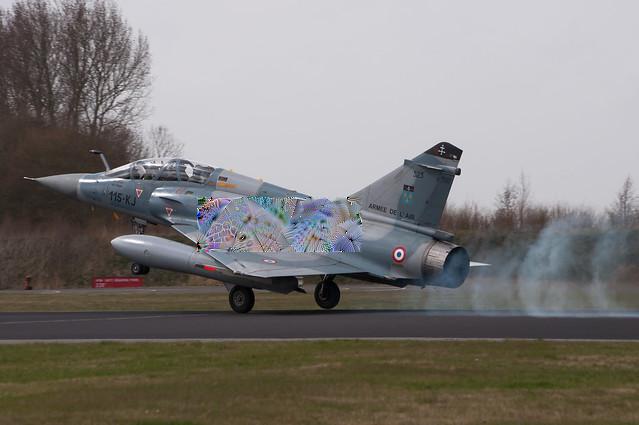}\\
\hline
Faster R-CNN prediction for image augmented with patch & \includegraphics[width=0.3\textwidth]{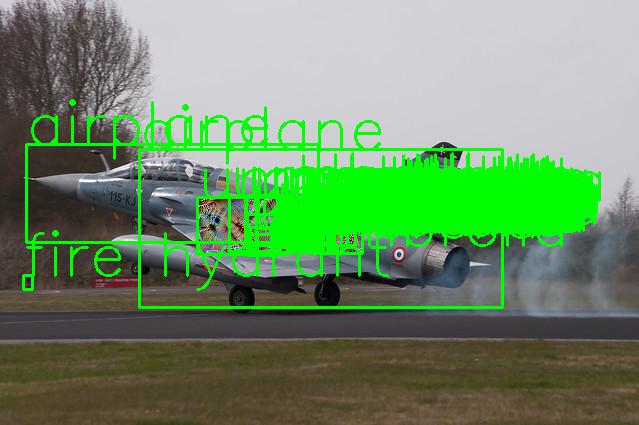} &
\includegraphics[width=0.3\textwidth]{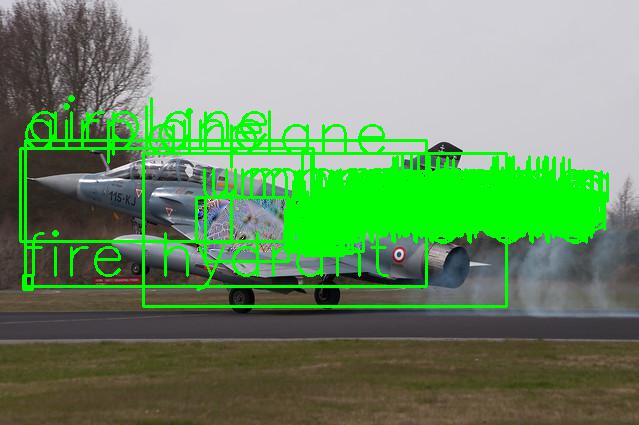} \\
\hline
\multicolumn{3}{l}{$^{\mathrm{a}}$reference Table~\ref{experiment_hyperparameters} for hyperparameters used}
\end{tabular}
\label{final_dpatch_vs_imperceptible_patch}
\end{center}
\end{table*}
\begin{table*}[t]
\caption{Experiment Hyperparameters}
\begin{center}
\begin{tabular}{|c|c|c|c|}
\hline
& \textbf{Initial Configuration} & \textbf{Robust-DPatch} & \textbf{Imperceptible Patch} \\
\hline
Iterations per Step & \multicolumn{3}{c|}{1000} \\
\hline
Deception Update Frequency controller ($n$) & \multicolumn{3}{c|}{1} \\
\hline
DLR & \multicolumn{3}{c|}{0.1} \\
\hline
Deception Update Momentum & \multicolumn{3}{c|}{0.9} \\
\hline
DLR Decay Amount & \multicolumn{3}{c|}{0.95} \\
\hline
DLR Decay Frequency & \multicolumn{3}{c|}{5 steps} \\
\hline
Number of Steps & 50 & 100 & 100\\
\hline
Maximum PLR & 0.5 & n.a. & 0.5 \\
\hline
Perceptibility Update Momentum  & 0.9 & n.a. & 0.9 \\
\hline
Maximum PLR Decay Amount  & 0.95 & n.a. & 0.95 \\
\hline
PLR Decay Frequency & 5 steps & n.a. & 5 steps \\
\hline
Inital Patch Configuration & Random & Random & Hybrid \\
\hline
\end{tabular}
\label{experiment_hyperparameters}
\end{center}
\end{table*}
\newpage
\section{Discussion \label{Discussion}}
The results demonstrate it is possible to create effective imperceptible patches, with the performance of the patches supporting our algorithm. Our method resulted in patches that are significantly less perceptible but as robust (based on mAP) as their Robust-DPatch counter parts. Such perceptibility differences are highlighted in Table \ref{final_dpatch_vs_imperceptible_patch}. 

 It is hypothesised that updates to our algorithm could produce patches with lower PerC distances, allowing such patch's presence to humans being further disguised. Borrowing from more traditional camouflage literature \cite{cuthill2019camouflage}, the future algorithm could incorporate more colours from the surrounding background into patches and not update the patch at random edge pixels to blur its outline. Inspired by \cite{brown2017adversarial}, the patch could also have standard aircraft markings superimposed on top.  

The positive experiment results could be further extended by:
\begin{itemize}
\item Using more images to show the more universal application of imperceptible patches and our proposed algorithm.
\item Utilising more than the prediction box to set patch size and location, aiming to ensure a patch will sit within an aircraft's bounds. 
\item Testing against a known and an unknown detection network. A black-box test is recommended as the specific computer vision algorithm implemented by adversaries will most likely be unknown for live use.
\item Executing ablation studies for 100 steps and the final comparison study for 200 steps, to allow for more perceptibility optimisation, given that the space is likely more complex as discussed in Section \ref{sec:Results}.
\end{itemize}
\section{Broader Impact \label{Broader_Impact}}
Benefits from this research may include methods that develop patches capable of defending against intrusive computer vision based recognition systems. Within a military context, these patches may assist in protecting personnel and aircraft from advanced tracking systems. A detection system trained with examples augmented with patches could become a more robust network. However, the algorithms and patches that have been developed may have negative impacts as they could provide felons and military insurgents with a potential method to disguise themselves from autonomous monitoring systems.  This could lead to more intrusive recognition systems. A malicious actor could use patches placed in a car window or on a sign, for example, to suppress automatic object detection by an autonomous car’s vision system (e.g., pedestrians, other cars, street signs). Alarmingly, if imperceptible robust patches are used, no object would need to be actively manipulated within the environment and being imperceptible, patches may therefore be difficult to police.
\section{Conclusions \label{Conclusions}}
Our research examined improvements to imperceptible adversarial patches designed to attack object detection algorithms. We designed a unique method to develop imperceptible patches capable of camouflaging large military assets from computer vision-enabled technologies. We created a robust imperceptible patch by combining a perceptibility loss function with the standard Robust-DPatch deception updates. The effectiveness of the produced patches against the Faster R-CNN detection algorithm when applied to aircraft images was investigated. Results indicated that our imperceptible patches can effectively camouflage an aircraft in an image from an object detection system. This shows the potential of the method to be used by the military to disguise large assets from modern computer vision enabled systems. Future work optimising the algorithm could result in more effective and imperceptible patches. This paper has proposed focuses for such future research and development. Continued study of adversarial techniques in computer vision and its idiosyncrasies will identify more resilient and universal classification and detection networks in addition to defending against them.
\section{Future Work \label{Future_Work}}
There are many directions for future research and development of imperceptible patches. Initially recommended is a focus on expanding the experiments to more general settings. Points of focus could include:
\begin{itemize}
\item Using more aircraft images, focusing on aircraft in a variety of environments. Using more training examples may also enable hyperparameters to be generalised. 
\item Investigating the effect and efficacy of image clipping after each deception and perceptibility update, which opposes the current paradigm of clipping at the end of each iteration. 
\item Further development of initial patch configurations and exploration of their effects on final patch robustness and perceptibility. Further work could also analyse the maximum and minimum RGB values for each channel in the image segment. The initial random noise patch could then be generated with RGB channels bounded by the calculated channel maximum and minimum.   
\item Further study and optimisation of image transformations. Though image rotations should potentially be limited to [-90, 90]° as an aircraft is realistically unlikely to be rotated outside of this range. Based on \cite{lee2019physical, brown2017adversarial, den2020adversarial, liu2018dpatch}, these transformations would then help to create more universal patches.
\item Researching the effect of using Project Gradient Descent (PGD) \cite{madry2017towards} to perform deception updates, as discussed in \cite{zhao2020towards}, FSGM yields poorer imperceptibility compared to PGD
\item Decaying the DLR using cosine annealing, such as \cite{zhao2020towards} performed. 
\end{itemize}

Future research could focus on using semantic segmentation \cite{long2015fully, hariharan2014simultaneous, gupta2014learning, pinheiro2014recurrent, girshick2014rich} to shape patches to the bounds of the aircraft. Using semantic patch bounding is a significant step towards more printable patches as once inside aircraft bounds, these patches could be used as an aircraft decal.

Further development should account for the plane’s contours when training and developing patches. Accounting for contours could allow patches to exploit shadowing effects whilst developing imperceptibility. Contoured patches may also be another vital training component in developing printable patches, as contouring may introduce distortions to the patch affecting performance.

A more comprehensive data set could be used to create more universal patches. An interesting experiment could test the efficacy of using several images of the same aircraft but from different angles. Thus, this paradigm would create a patch universal to a specific aircraft model. These experiments could further investigate different patches for different locations on the aircraft. For example, patches for the aircraft's underbelly could be trained only on images of that aircraft in the sky as it is unlikely that vision technology would see a grounded aircraft's underbelly. However, patches for the aircraft's top should be trained using grounded and flying aircraft examples. A mixture of examples is suggested as a drone or missile will likely see the top of a grounded and flying aircraft.  

Future work could also investigate the effect that multiple aircraft, and therefore patches, have on detection accuracy. Evidence suggests that Robust-DPatch \cite{lee2019physical} can suppress detections even when the patch is far from an object. Consequently, perhaps only one patch is needed for a swarm of planes to suppress detections effectively. Further, perhaps multiple patched planes could be highly effective as evidence suggests the effectiveness of multiple patches \cite{den2020adversarial}.

Targeted cases could be investigated further, such as using a person as the adversarial target. Use of specific targets may be necessary as tracking tools could be programmed to still target out of context detections. These tools may be programmed this way, as seeing an out-of-context object indicates the presence of an adversarial patch. However, according to the Geneva conventions, ejected pilots cannot be targeted \cite{genevaconventions} and consequently, one assumes that missiles targeting planes will be programmed not to target humans as a detected human is likely an ejected pilot. Therefore, using a person as an adversarial target may present a scheme to exploit detection algorithms.

Targeted cases could also investigate using an aircraft as the adversarial target itself. These patches are then countermeasures and, like chaff, \cite{butters1982chaff}, could be dropped from an aircraft to attract missiles away. Further, specific plane models could be targeted. Specifically, a network would misclassify a more threatening aircraft as a more mundane counterpart, which may also help to mislead any human operators receiving data from these vision technologies. 

GAN generated patches \cite{liu2019perceptual, zhao2020ap} could also incorporate perceptibility calculations into the generator’s loss function. This paradigm could further benefit from producing a discriminator network resilient to imperceptible adversarial patches.
\bibliographystyle{plain}
\bibliography{references}

\begin{thebibliography}{10}

\bibitem{aurdal2019adversarial}
Lars Aurdal, Kristin~Hammarstr{\o}m L{\o}kken, Runhild~Aae Klausen, Alvin
  Brattli, and Hans~Christian Palm.
\newblock Adversarial camouflage for naval vessels.
\newblock In {\em Artificial Intelligence and Machine Learning in Defense
  Applications}, volume 11169, page 111690K. International Society for Optics
  and Photonics, 2019.

\bibitem{belmonte2019computer}
Lidia~Mar{\'\i}a Belmonte, Rafael Morales, and Antonio Fern{\'a}ndez-Caballero.
\newblock Computer vision in autonomous unmanned aerial vehicles - a systematic
  mapping study.
\newblock {\em Applied Sciences}, 9(15):3196, 2019.

\bibitem{berker2017feature}
K~Berker~Logoglu, Hazal Lezki, M~Kerim~Yucel, Ahu Ozturk, Alper Kucukkomurler,
  Batuhan Karagoz, Erkut Erdem, and Aykut Erdem.
\newblock Feature-based efficient moving object detection for low-altitude
  aerial platforms.
\newblock In {\em Proceedings of the IEEE International Conference on Computer
  Vision Workshops}, pages 2119--2128, 2017.

\bibitem{bose2018adversarial}
Avishek~Joey Bose and Parham Aarabi.
\newblock Adversarial attacks on face detectors using neural net based
  constrained optimization.
\newblock In {\em 2018 IEEE 20th International Workshop on Multimedia Signal
  Processing (MMSP)}, pages 1--6. IEEE, 2018.

\bibitem{brown2017adversarial}
Tom~B Brown, Dandelion Man{\'e}, Aurko Roy, Mart{\'\i}n Abadi, and Justin
  Gilmer.
\newblock Adversarial patch.
\newblock {\em arXiv preprint arXiv:1712.09665}, 2017.

\bibitem{butters1982chaff}
Brian~CF Butters.
\newblock Chaff.
\newblock In {\em IEE Proceedings F-Communications, Radar and Signal
  Processing}, volume 129, pages 197--201. IET, 1982.

\bibitem{carlini2017towards}
Nicholas Carlini and David Wagner.
\newblock Towards evaluating the robustness of neural networks.
\newblock In {\em 2017 ieee symposium on security and privacy (sp)}, pages
  39--57. IEEE, 2017.

\bibitem{chen2018shapeshifter}
Shang-Tse Chen, Cory Cornelius, Jason Martin, and Duen Horng~Polo Chau.
\newblock Shapeshifter: Robust physical adversarial attack on faster r-cnn
  object detector.
\newblock In {\em Joint European Conference on Machine Learning and Knowledge
  Discovery in Databases}, pages 52--68. Springer, 2018.

\bibitem{cheung2020graph}
Mark Cheung, John Shi, Oren Wright, Lavendar~Y Jiang, Xujin Liu, and
  Jos{\'e}~MF Moura.
\newblock Graph signal processing and deep learning: Convolution, pooling, and
  topology.
\newblock {\em IEEE Signal Processing Magazine}, 37(6):139--149, 2020.

\bibitem{cuthill2019camouflage}
IC~Cuthill.
\newblock Camouflage.
\newblock {\em Journal of Zoology}, 308(2):75--92, 2019.

\bibitem{den2020adversarial}
Richard den Hollander, Ajaya Adhikari, Ioannis Tolios, Michael van Bekkum,
  Anneloes Bal, Stijn Hendriks, Maarten Kruithof, Dennis Gross, Nils Jansen,
  Guillermo Perez, et~al.
\newblock Adversarial patch camouflage against aerial detection.
\newblock In {\em Artificial Intelligence and Machine Learning in Defense
  Applications II}, volume 11543, page 115430F. International Society for
  Optics and Photonics, 2020.

\bibitem{eykholt2018robust}
Kevin Eykholt, Ivan Evtimov, Earlence Fernandes, Bo~Li, Amir Rahmati, Chaowei
  Xiao, Atul Prakash, Tadayoshi Kohno, and Dawn Song.
\newblock Robust physical-world attacks on deep learning visual classification.
\newblock In {\em Proceedings of the IEEE conference on computer vision and
  pattern recognition}, pages 1625--1634, 2018.

\bibitem{girshick2014rich}
Ross Girshick, Jeff Donahue, Trevor Darrell, and Jitendra Malik.
\newblock Rich feature hierarchies for accurate object detection and semantic
  segmentation.
\newblock In {\em Proceedings of the IEEE conference on computer vision and
  pattern recognition}, pages 580--587, 2014.

\bibitem{goodfellow2014explaining}
Ian~J Goodfellow, Jonathon Shlens, and Christian Szegedy.
\newblock Explaining and harnessing adversarial examples.
\newblock {\em arXiv preprint arXiv:1412.6572}, 2014.

\bibitem{gu2018recent}
Jiuxiang Gu, Zhenhua Wang, Jason Kuen, Lianyang Ma, Amir Shahroudy, Bing Shuai,
  Ting Liu, Xingxing Wang, Gang Wang, Jianfei Cai, et~al.
\newblock Recent advances in convolutional neural networks.
\newblock {\em Pattern Recognition}, 77:354--377, 2018.

\bibitem{gupta2014learning}
Saurabh Gupta, Ross Girshick, Pablo Arbel{\'a}ez, and Jitendra Malik.
\newblock Learning rich features from rgb-d images for object detection and
  segmentation.
\newblock In {\em European conference on computer vision}, pages 345--360.
  Springer, 2014.

\bibitem{hariharan2014simultaneous}
Bharath Hariharan, Pablo Arbel{\'a}ez, Ross Girshick, and Jitendra Malik.
\newblock Simultaneous detection and segmentation.
\newblock In {\em European conference on computer vision}, pages 297--312.
  Springer, 2014.

\bibitem{kechagias20163d}
Odysseas Kechagias-Stamatis, Nabil Aouf, and Mark~A Richardson.
\newblock 3d automatic target recognition for future lidar missiles.
\newblock {\em IEEE Transactions on Aerospace and Electronic systems},
  52(6):2662--2675, 2016.

\bibitem{krizhevsky2012imagenet}
Alex Krizhevsky, Ilya Sutskever, and Geoffrey~E Hinton.
\newblock Imagenet classification with deep convolutional neural networks.
\newblock {\em Advances in neural information processing systems},
  25:1097--1105, 2012.

\bibitem{lee2019physical}
Mark Lee and Zico Kolter.
\newblock On physical adversarial patches for object detection.
\newblock {\em arXiv preprint arXiv:1906.11897}, 2019.

\bibitem{li2020deep}
Shan Li and Weihong Deng.
\newblock Deep facial expression recognition: A survey.
\newblock {\em IEEE transactions on affective computing}, 2020.

\bibitem{lin2014microsoft}
Tsung-Yi Lin, Michael Maire, Serge Belongie, James Hays, Pietro Perona, Deva
  Ramanan, Piotr Doll{\'a}r, and C~Lawrence Zitnick.
\newblock Microsoft coco: Common objects in context.
\newblock In {\em European conference on computer vision}, pages 740--755.
  Springer, 2014.

\bibitem{liu2019perceptual}
Aishan Liu, Xianglong Liu, Jiaxin Fan, Yuqing Ma, Anlan Zhang, Huiyuan Xie, and
  Dacheng Tao.
\newblock Perceptual-sensitive gan for generating adversarial patches.
\newblock In {\em Proceedings of the AAAI conference on artificial
  intelligence}, volume~33, pages 1028--1035, 2019.

\bibitem{liu2018dpatch}
Xin Liu, Huanrui Yang, Ziwei Liu, Linghao Song, Hai Li, and Yiran Chen.
\newblock Dpatch: An adversarial patch attack on object detectors.
\newblock {\em arXiv preprint arXiv:1806.02299}, 2018.

\bibitem{long2015fully}
Jonathan Long, Evan Shelhamer, and Trevor Darrell.
\newblock Fully convolutional networks for semantic segmentation.
\newblock In {\em Proceedings of the IEEE conference on computer vision and
  pattern recognition}, pages 3431--3440, 2015.

\bibitem{luo2001development}
M~Ronnier Luo, Guihua Cui, and Bryan Rigg.
\newblock The development of the cie 2000 colour-difference formula: Ciede2000.
\newblock {\em Color Research \& Application: Endorsed by Inter-Society Color
  Council, The Colour Group (Great Britain), Canadian Society for Color, Color
  Science Association of Japan, Dutch Society for the Study of Color, The
  Swedish Colour Centre Foundation, Colour Society of Australia, Centre
  Fran{\c{c}}ais de la Couleur}, 26(5):340--350, 2001.

\bibitem{madry2017towards}
Aleksander Madry, Aleksandar Makelov, Ludwig Schmidt, Dimitris Tsipras, and
  Adrian Vladu.
\newblock Towards deep learning models resistant to adversarial attacks.
\newblock {\em arXiv preprint arXiv:1706.06083}, 2017.

\bibitem{masi2018deep}
Iacopo Masi, Yue Wu, Tal Hassner, and Prem Natarajan.
\newblock Deep face recognition: A survey.
\newblock In {\em 2018 31st SIBGRAPI conference on graphics, patterns and
  images (SIBGRAPI)}, pages 471--478. IEEE, 2018.

\bibitem{mazurowski2019deep}
Maciej~A Mazurowski, Mateusz Buda, Ashirbani Saha, and Mustafa~R Bashir.
\newblock Deep learning in radiology: An overview of the concepts and a survey
  of the state of the art with focus on mri.
\newblock {\em Journal of magnetic resonance imaging}, 49(4):939--954, 2019.

\bibitem{moosavi2017universal}
Seyed-Mohsen Moosavi-Dezfooli, Alhussein Fawzi, Omar Fawzi, and Pascal
  Frossard.
\newblock Universal adversarial perturbations.
\newblock In {\em Proceedings of the IEEE conference on computer vision and
  pattern recognition}, pages 1765--1773, 2017.

\bibitem{nicolae2018adversarial}
Maria-Irina Nicolae, Mathieu Sinn, Minh~Ngoc Tran, Beat Buesser, Ambrish Rawat,
  Martin Wistuba, Valentina Zantedeschi, Nathalie Baracaldo, Bryant Chen, Heiko
  Ludwig, et~al.
\newblock Adversarial robustness toolbox v1. 0.0.
\newblock {\em arXiv preprint arXiv:1807.01069}, 2018.

\bibitem{genevaconventions}
International~Committee of~the Red~Cross.
\newblock Protocol additional to the geneva conventions of 12 august 1949, and
  relating to the protection of victims of international armed conflicts
  (protocol i), 8 june 1977.

\bibitem{papernot2016limitations}
Nicolas Papernot, Patrick McDaniel, Somesh Jha, Matt Fredrikson, Z~Berkay
  Celik, and Ananthram Swami.
\newblock The limitations of deep learning in adversarial settings.
\newblock In {\em 2016 IEEE European symposium on security and privacy
  (EuroS\&P)}, pages 372--387. IEEE, 2016.

\bibitem{pinheiro2014recurrent}
Pedro Pinheiro and Ronan Collobert.
\newblock Recurrent convolutional neural networks for scene labeling.
\newblock In {\em International conference on machine learning}, pages 82--90.
  PMLR, 2014.

\bibitem{rawat2017deep}
Waseem Rawat and Zenghui Wang.
\newblock Deep convolutional neural networks for image classification: A
  comprehensive review.
\newblock {\em Neural computation}, 29(9):2352--2449, 2017.

\bibitem{ren2015faster}
Shaoqing Ren, Kaiming He, Ross Girshick, and Jian Sun.
\newblock Faster r-cnn: Towards real-time object detection with region proposal
  networks.
\newblock {\em Advances in neural information processing systems}, 28:91--99,
  2015.

\bibitem{rony2019decoupling}
J{\'e}r{\^o}me Rony, Luiz~G Hafemann, Luiz~S Oliveira, Ismail~Ben Ayed, Robert
  Sabourin, and Eric Granger.
\newblock Decoupling direction and norm for efficient gradient-based l2
  adversarial attacks and defenses.
\newblock In {\em Proceedings of the IEEE/CVF Conference on Computer Vision and
  Pattern Recognition}, pages 4322--4330, 2019.

\bibitem{sharif2016accessorize}
Mahmood Sharif, Sruti Bhagavatula, Lujo Bauer, and Michael~K Reiter.
\newblock Accessorize to a crime: Real and stealthy attacks on state-of-the-art
  face recognition.
\newblock In {\em Proceedings of the 2016 acm sigsac conference on computer and
  communications security}, pages 1528--1540, 2016.

\bibitem{sharif2019general}
Mahmood Sharif, Sruti Bhagavatula, Lujo Bauer, and Michael~K Reiter.
\newblock A general framework for adversarial examples with objectives.
\newblock {\em ACM Transactions on Privacy and Security (TOPS)}, 22(3):1--30,
  2019.

\bibitem{su2019one}
Jiawei Su, Danilo~Vasconcellos Vargas, and Kouichi Sakurai.
\newblock One pixel attack for fooling deep neural networks.
\newblock {\em IEEE Transactions on Evolutionary Computation}, 23(5):828--841,
  2019.

\bibitem{szegedy2013intriguing}
Christian Szegedy, Wojciech Zaremba, Ilya Sutskever, Joan Bruna, Dumitru Erhan,
  Ian Goodfellow, and Rob Fergus.
\newblock Intriguing properties of neural networks.
\newblock {\em arXiv preprint arXiv:1312.6199}, 2013.

\bibitem{thys2019fooling}
Simen Thys, Wiebe Van~Ranst, and Toon Goedem{\'e}.
\newblock Fooling automated surveillance cameras: adversarial patches to attack
  person detection.
\newblock In {\em Proceedings of the IEEE/CVF Conference on Computer Vision and
  Pattern Recognition Workshops}, pages 0--0, 2019.

\bibitem{artTrustedAI}
Trusted-AI.
\newblock Trusted-ai/adversarial-robustness-toolbox: Adversarial robustness
  toolbox (art) - python library for machine learning security - evasion,
  poisoning, extraction, inference - red and blue teams.
\newblock https://github.com/Trusted-AI/adversarial-robustness-toolbox.

\bibitem{wang2021object}
Shaobo Wang, Cheng Zhang, Di~Su, and Tianqi Sun.
\newblock A object detection method for missile-borne images based on improved
  yolov3.
\newblock In {\em Journal of Physics: Conference Series}, volume 1880, page
  012018. IOP Publishing, 2021.

\bibitem{wu2020making}
Zuxuan Wu, Ser-Nam Lim, Larry~S Davis, and Tom Goldstein.
\newblock Making an invisibility cloak: Real world adversarial attacks on
  object detectors.
\newblock In {\em European Conference on Computer Vision}, pages 1--17.
  Springer, 2020.

\bibitem{zhao2020ap}
Guoping Zhao, Mingyu Zhang, Jiajun Liu, Yaxian Li, and Ji-Rong Wen.
\newblock Ap-gan: Adversarial patch attack on content-based image retrieval
  systems.
\newblock {\em GeoInformatica}, pages 1--31, 2020.

\bibitem{zhao2020towards}
Zhengyu Zhao, Zhuoran Liu, and Martha Larson.
\newblock Towards large yet imperceptible adversarial image perturbations with
  perceptual color distance.
\newblock In {\em Proceedings of the IEEE/CVF Conference on Computer Vision and
  Pattern Recognition}, pages 1039--1048, 2020.

\bibitem{zhu2018visdrone}
Pengfei Zhu, Longyin Wen, Dawei Du, Xiao Bian, Haibin Ling, Qinghua Hu, Qinqin
  Nie, Hao Cheng, Chenfeng Liu, Xiaoyu Liu, et~al.
\newblock Visdrone-det2018: The vision meets drone object detection in image
  challenge results.
\newblock In {\em Proceedings of the European Conference on Computer Vision
  (ECCV) Workshops}, pages 0--0, 2018.

\end{thebibliography}
\end{document}